\documentclass[11pt,a4paper]{article}
\usepackage[hyperref]{acl2021}
\usepackage{times}
\usepackage{latexsym}

\usepackage{microtype}

\usepackage{graphicx}
\usepackage{latexsym}
\usepackage{epstopdf}
\usepackage{amssymb}
\usepackage{amsmath}
\usepackage{multirow}
\usepackage{xspace}
\usepackage{enumitem}
\usepackage{pifont}

\newcommand{\our}{\mbox{CIL}\xspace}
\newcommand{\eg}{\textit{e.g.}}
\newcommand{\ie}{\textit{i.e.}}

\aclfinalcopy 

\title{CIL: Contrastive Instance Learning Framework for Distantly Supervised Relation Extraction}

\author {
    Tao Chen\textsuperscript{\rm 1,2},
    Haizhou Shi\textsuperscript{\rm 1},
    Siliang Tang\textsuperscript{\rm 1,2}\thanks{\quad Corresponding author} ,
    Zhigang Chen\textsuperscript{\rm 3}, \\
    \textbf{Fei Wu}\textsuperscript{\rm 1} \textbf{\&}
    \textbf{Yueting Zhuang}\textsuperscript{\rm 1}
    \\
    \textsuperscript{\rm 1}Zhejiang University \\
    \textsuperscript{\rm 2}Alibaba-Zhejiang University Joint Research Institute of Frontier Technologies \\
    \textsuperscript{\rm 3}State Key Laboratory of Cognitive Intelligence, Hefei, China \\
    \texttt{\{ttc, shihaizhou, siliang, yzhuang\}@zju.edu.cn}\\
    \texttt{zgchen@iflytek.com, wufei@cs.zju.edu.cn}
}

\date{}

\begin{document}
\maketitle

\begin{abstract}

The journey of reducing noise from distant supervision (DS) generated training data has been started since the DS was first introduced into the relation extraction (RE) task. 
For the past decade, researchers apply the multi-instance learning (MIL) framework to find the most reliable feature from a bag of sentences.
Although the pattern of MIL bags can greatly reduce DS noise, it fails to represent many other useful sentence features in the datasets. 
In many cases, these sentence features can only be acquired by extra sentence-level human annotation with heavy costs.
Therefore, the performance of distantly supervised RE models is bounded.
In this paper, we go beyond typical MIL framework and propose a novel \underline{C}ontrastive \underline{I}nstance \underline{L}earning (CIL) framework.
Specifically, we regard the initial MIL as the relational triple encoder and constraint positive pairs against negative pairs for each instance.
Experiments demonstrate the effectiveness of our proposed framework, with significant improvements over the previous methods on NYT10, GDS and KBP.

\end{abstract}

\section{Introduction}
\label{intro}

Relation extraction (RE) aims at predicting the relation between entities based on their context.
Several studies have been carried out to handle this crucial and complicated task over decades as the extracted information can serve as a significant role for many downstream tasks. 
Since the amount of training data generally limits traditional supervised RE systems, current RE systems usually resort to distant supervision (DS) to fetch abundant training data by aligning knowledge bases (KBs) and texts.
However, such a heuristic way inevitably introduces some noise to the generated data.
Training a robust and unbiased RE system under DS data noise becomes the biggest challenge for distantly supervised relation extraction (DSRE). 

\begin{figure}[t]
\centerline{\includegraphics[scale=0.65]{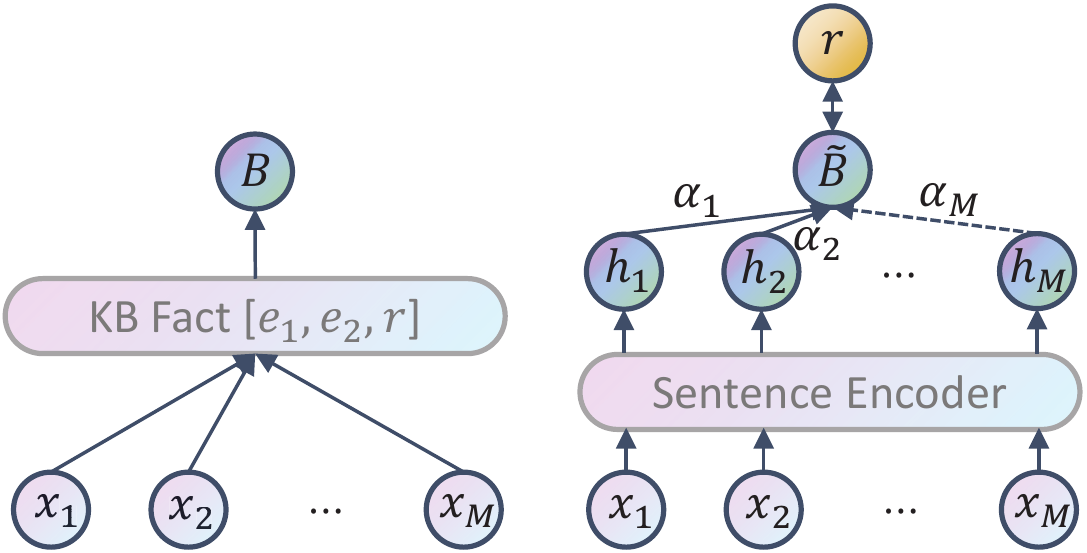}}
\caption{\label{bag re} Classical MIL framework for DSRE.
(Left) A set of instances ($x_1,x_2,\dots,x_m$) with the same KB fact [$e_1,e_2,r$] form a bag $B$;
(Right) The MIL framework trains the DSRE model at bag level(\small$\widetilde B:\sum \alpha_i h_i$).}
\end{figure}

With awareness of the existing DS noise, ~\citet{zeng2015distant} introduces the multi-instance learning (MIL) framework to DSRE by dividing training instances into several bags and using bags as new data units.
Regarding the strategy for selecting instances inside the bag, the soft attention mechanism proposed by~\citet{lin2016neural} is widely used for its better performance than the hard selection method.
The ability to form accurate representations from noisy data makes the MIL framework soon become a paradigm of following-up works.

However, we argue that the MIL framework is effective to alleviate data noise for DSRE, but is not data-efficient indeed:
As Figure~\ref{bag re} shows: The attention mechanism in the MIL can help select relatively informative instances~(\eg $h_1,h_2$) inside the bag, but may ignore the potential information of other abundant instances~(\eg $h_m$).
In other words, no matter how many instances a bag contains, only the formed bag-level representation can be used for further training in the MIL, which is quite inefficient.
Thus, our focus is on how to make the initial MIL framework \textit{efficient enough to leverage all instances} while \textit{maintaining the ability to obtain an accurate model under DS data noise}?

Here, we propose a contrastive-based method to help the MIL framework learn efficiently.
In detail, we regard the initial MIL framework as the bag encoder, which provides relatively accurate representations for different relational triples.
Then we develop contrastive instance learning (\our) to utilize each instance in an unsupervised manner:
In short, the goal of our \our is that the instances sharing the same relational triples (\ie positive pairs) ought to be close in the semantic space, while the representations of instances with different relational triples (\ie negative pairs) should be far away. 

Experiments on three public DSRE benchmarks --- NYT10~\citep{riedel2010modeling,hoffmann2011knowledge}, GDS~\citep{jat2018improving} and KBP~\citep{ling2012fine} demonstrate the effectiveness of our proposed framework \our, with consistent improvements over several baseline models and far exceed the state-of-the-art (SOTA) systems. Furthermore, the ablation study shows the rationality of our proposed positive/negative pair construction strategy.

Accordingly, the major contributions of this paper are summarized as follows:

\begin{itemize}
    \item We discuss the long-standing MIL framework and point out that it can not effectively utilize abundant instances inside MIL bags.
    \item We propose a novel contrastive instance learning method to boost the DSRE model performances under the MIL framework.
    \item Evaluation on held-out and human-annotated sets shows that \our leads to significant improvements over the previous SOTA models. 
\end{itemize}

\section{Methodology}

In this paper, we argue that the MIL framework is effective to denoise but is not efficient enough, as the initial MIL framework only leverages the formed bag-level representations to train models and sacrifices the potential information of numerous instances inside bags.
Here, we go beyond the typical MIL framework and develop a novel contrastive instance learning framework to solve the above issue, which can prompt DSRE models to utilize each instance.
A formal description of our proposed \our framework is illustrated as follows.

\subsection{Input Embeddings}
\label{sec: inputs}

\paragraph{Token Embedding} 
For input sentence/instance $x$, we utilize BERT Tokenizer to split it into several tokens: ($t_1,t_2,\dots e_1\dots e_2 \dots t_L$), where $e_1, e_2$ are the tokens corresponding to the two entities, and $L$ is the max length of all input sequences.
Following standard practices~\citep{devlin-etal-2019-bert}, we add two special tokens to mark the beginning ([CLS]) and the end ([SEP]) of sentences. 

In BERT, token [CLS] typically acts as a pooling token representing the whole sequence for downstream tasks.
However, this pooling representation considers entity tokens $e_1$ and $e_2$ as equivalent to other common word tokens $t_i$, which has been proven~\citep{baldini-soares-etal-2019-matching} to be unsuitable for RE tasks.
To encode the sentence in an entity-aware manner, we add four extra special tokens ([H-CLS], [H-SEP]) and ([T-CLS], [T-SEP]) to mark the beginning and the end of two entities.

\paragraph{Position Embedding}

In the Transformer attention mechanism~\citep{vaswani2017attention}, positional encodings are injected to make use of the order of the sequence.
Precisely, the learned position embedding has the same dimension as the token embedding so that the two can be summed.

\begin{figure}[htbp]
\centerline{\includegraphics[scale=0.65]{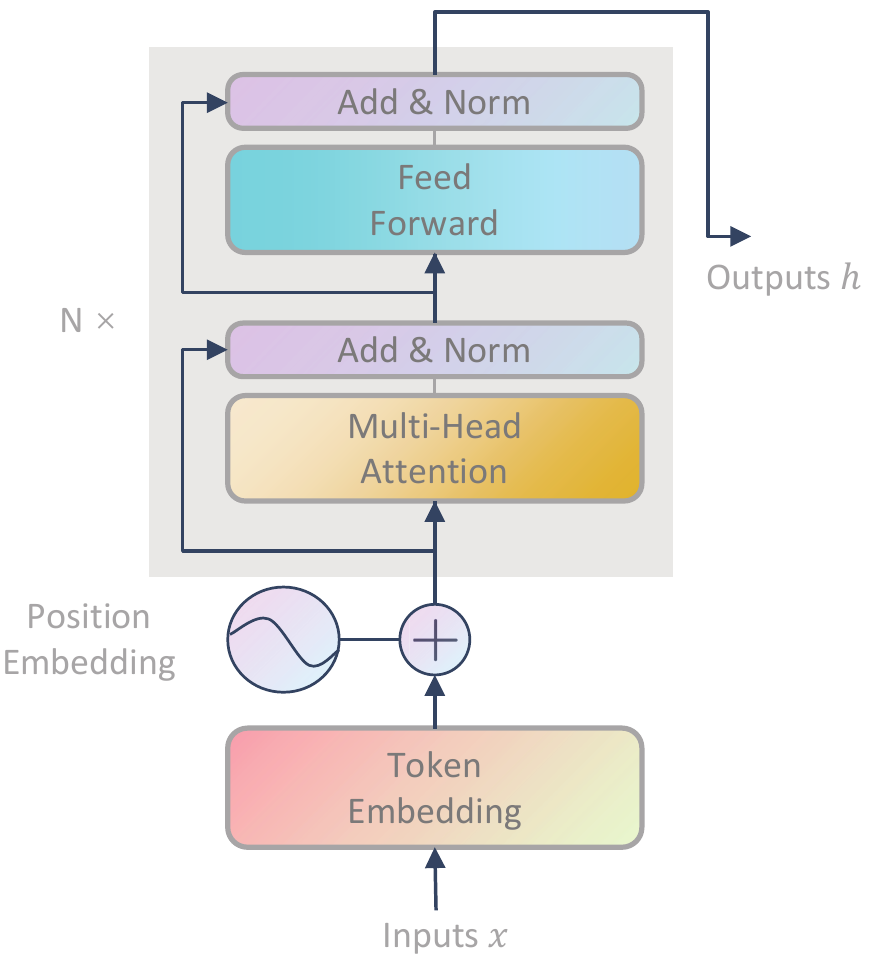}}
\caption{BERT Encoder: N × Transformer Blocks.}
\label{transformer block}
\end{figure}

\subsection{Sentence Encoder}
\label{sec: sentence encoder}

BERT Encoder (Transformer Blocks, see Figure~\ref{transformer block}) transforms the above embedding inputs (token embedding \& position embedding) into hidden feature vectors: ($h_1,h_2, \dots h_{e_1} \dots h_{e_2} \dots h_L)$, where $h_{e_1}$ and $h_{e_2}$ are the feature vectors corresponding to the entities $e_1$ and $e_2$.
By concatenating the two entity hidden vectors, we can obtain the entity-aware sentence representation $h=[h_{e_1};h_{e_2}]$ for the input sequence $x$. We denote the sentence encoder $\mathcal{H}$ as:
\begin{equation*}
    \mathcal{H}(x)=[h_{e_1};h_{e_2}]=h
\end{equation*}

\subsection{Bag Encoder}
\label{sec:bag encoder}

Under the MIL framework, a couple of instances $x$ with the same relational triple $[e_1,e_2,r]$ form a bag $B$.
We aim to design a bag encoder $\mathcal{F}$ to obtain representation $\widetilde B$ for bag $B$, and the obtained bag representation is also a representative of the current relational triple $[e_1, e_2, r]$, which is defined as:
\begin{equation*}
    \mathcal{F}(B)=\mathcal{F}([e_1,e_2,r])=\widetilde B
\end{equation*}

With the help of the sentence encoder described in section~\ref{sec: sentence encoder}, each instance $x_i$ in bag $B$ can be first encoded to its entity-aware sentence representation $h_i=\mathcal{H}(x_i)$.
Then the bag representation $\widetilde B$ can be regarded as an aggregation of all instances' representations, which is further defined as:
\begin{equation*}
    \mathcal{F}([e_1,e_2,r])=\widetilde B=\sum_{i=1}^{K} \alpha_i h_i
\end{equation*}
where $K$ is the bag size. As for the choice of weight $\alpha_i$, we follow the soft attention mechanism used in~\citep{lin2016neural}, where $\alpha_i$ is the normalized attention score calculated by a query-based function $f_i$ that measures how well the sentence representation $h_i$ and the predict relation $r$ matches:
\begin{equation*}
    \alpha_i=\frac{e^{f_i}}{\sum_j e^{f_j}}
\end{equation*}
where $f_i=h_iA q_r$, $A$ is a weighted diagonal matrix and $q_r$ is
the query vector which indicates the representation of relation $r$ (randomly initialized).

Then, to train such a bag encoder parameterized by $\theta$, a simple fully-connected layer with activation function $\textit{softmax}$ is added to map the hidden feature vector $\widetilde B$ to a conditional probability distribution $p(r|\widetilde B,\theta)$, and this can be defined as:
\begin{equation*}
    p(r|\widetilde B,\theta)=\frac{e^{o_r}}{\sum_{i=1}^{n_r} e^{o_i}}
\end{equation*}
where $o=M\widetilde B+b$ is the score associated to all relation types, $n_r$ is the total number of relations, $M$ is a projection matrix, and $b$ is the bias term.

And we define the objective of bag encoder using cross-entropy function as follows:
\begin{equation*}
    \mathcal{L_B}(\theta) = -\sum_{i=1} \log p(r_i|\widetilde B_i,\theta)
\end{equation*}

\subsection{Contrastive Instance Learning}
\label{sec: cil}

As illustrated in section~\ref{intro}, the goal of our framework \our is that the instances containing the same relational triples (\ie positive pairs) should be as close (\ie $\sim$) as possible in the hidden semantic space, and the instances containing different relational triples (\ie negative pairs) should be as far (\ie $\nsim$) away as possible in the space. A formal description is as follows.

Assume there is a batch bag input (with a batch size $G$): $(B_1,B_2,\dots,B_G)$, the relational triples of all bags are different from each other.
Each bag $B$ in the batch is constructed by a certain relational triple $[e_{1},e_{2},r]$, and all instances $x$ inside the bag satisfy this triple. The representation of the triple can be obtained by bag encoder as $\widetilde B$.

We pick any two bags $B_s$ and $B_{t:t\neq s}$ in the batch to further illustrate the process of contrastive instance learning. $B_s$ is defined as the source bag constructed with relational triple $[e_{s1}, e_{s2}, r_s]$ while $B_t$ is the target bag constructed with triple $[e_{t1}, e_{t2}, r_t]$. And we discuss the positive pair instance and negative pair instances for any instance $x_{s}$ in bag $B_s$.

It is worth noting that all bags are constructed automatically by the distantly supervised method, which extracts relational triples from instances in a heuristic manner and may introduce true/false positive label noise to the generated data.
In other words, though the instance $x$ is included in the bag with relational triple $[e_1,e_2,r]$, it may be noisy and fail to express the relation $r$.

\subsubsection{Positive Pair Construction}

\paragraph{Instance $x_s$  $\sim$ Random Instance $x_{s'}$}

One intuitive choice of selecting positive pair instance for instance $x_s$ is just picking another instance $x_{s'} \neq x_s$ from the bag $B$ randomly.  
However, both of the instances $x_s$ and $x_{s'}$ may suffer from data noise, and they are hard to express the same relational triple simultaneously.
Thus, taking instance $x_s$ and randomly selected instance $x_{s'}$ as a positive pair is not an optimal option.

\begin{figure}[htbp]
\centerline{\includegraphics[scale=0.65]{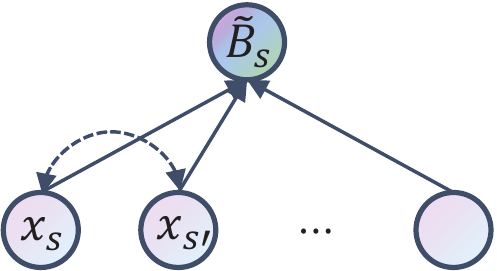}}
\caption{Instance $x_s$ $\sim$ Random Instance $x_{s'}$}
\end{figure}

\paragraph{Instance $x_s$ $\sim$ Relational Triple $\widetilde B_s$}

Another positive pair instance candidate for instance $x_s$ is the relational triple representation $\widetilde B_s$ of current bag $B$.
Though $\widetilde B_s$ can be regarded as a de-noised representation, $x_s$ may be still noisy and express other relation $r\neq r_s$.
Besides, the quality of constructed positive pairs heavily relies on the model performance of the bag encoder.

\begin{figure}[htbp]
\centerline{\includegraphics[scale=0.65]{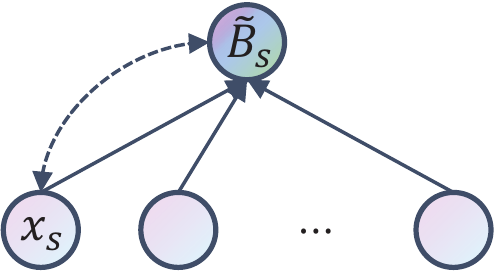}}
\caption{Instance $x_s$ $\sim$ Relational Triple $\widetilde B_s$}
\end{figure}

\paragraph{Instance $x_s$ $\sim$ Augmented Instance $x_s^{*}$}

From the above analysis, we can see that the general positive pair construction methods often encounter the challenge of DS noise. 
Here, we propose a noise-free positive pair construction method based on \textit{TF-IDF} data augmentation: If we only make small and controllable data augmentation to the original instance $x_s$, the augmented instance $x_s^*$ should satisfy the same relational triple with instance $x_s$.

\begin{figure}[htbp]
\centerline{\includegraphics[scale=0.65]{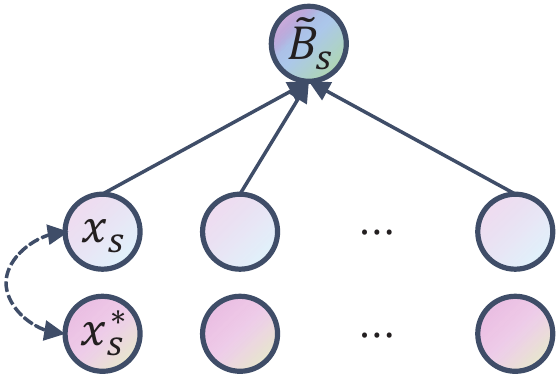}}
\caption{Instance $x_s$ $\sim$ Augmented Instance $x_s^*$}
\end{figure}

\begin{figure}[htbp]
\centerline{\includegraphics[scale=0.75]{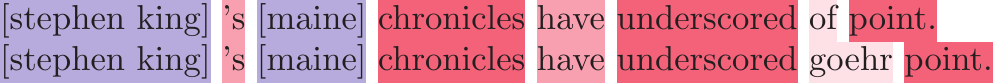}}
\caption{An example of word substitution: The low-scoring word \textit{of} is replaced with word \textit{goehr}, and entity words \textit{stephen king} and \textit{maine} are protected.} 
\label{tf-idf}
\end{figure}

In detail: 
(1) We first view each instance as a document and view each word in the instances as a term, then we train a \textit{TF-IDF} model on the total training corpus.
(2) Based on the trained \textit{TF-IDF} model, we insert/substitute some unimportant (low \textit{TF-IDF} score, see Figure~\ref{tf-idf}) words to/in instance $x_s$ with a specific ratio, and can obtain its augmented instance $x_s^*$. Particularly, special masks are added to entity words to avoid them being substituted.

\subsubsection{Negative Pair Construction}

\paragraph{Instance $x_s$ $\nsim$ Random Instance $x_{t}$}

Similarly, for instance $x_s$ in bag $B_s$, we can randomly select an instance $x_t$ from another different bag $B_t$ as its negative pair instance. 
Under this strategy, $x_s$ is far away from the average representation $\sum_{i=1}^K \alpha_i h_i$ of the bag $B_t$, where all $\alpha_i = \frac{1}{K}$ approximately.
And the randomly selected instance $x_t$ may be too noisy to represent the relational triple of bag $B_t$, so that the model performance may be influenced.

\begin{figure}[htbp]
\centerline{\includegraphics[scale=0.65]{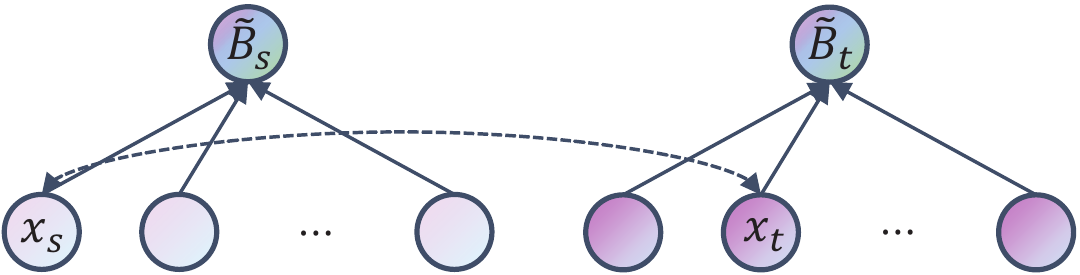}}
\caption{Instance $x_s$ $\nsim$ Random Instance $x_t$}
\end{figure}

\paragraph{Instance $x_s$ $\nsim$ Relational Triple $\widetilde B_t$}

Compared to the random selection strategy, using relational triple representation $\widetilde B_t$ as the negative pair instance for $x_s$ is a better choice to reduce the impact of data noise.
As the instance $x_i$ can be seen as be far away from a weighted representation $\sum_{i=1}^K \alpha_i h_i$ of the bag $B_t$, where all $\alpha_i$ are learnable.
Though the instance $x_s$ may still be noisy, $x_s$ and $\widetilde B_t$ can not belong to the same relational triple.

\begin{figure}[htbp]
\centerline{\includegraphics[scale=0.65]{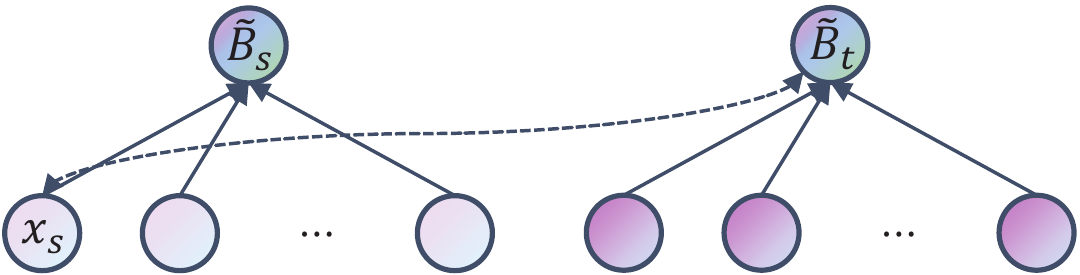}}
\caption{Instance $x_s$ $\nsim$ Relational Triple $\widetilde B_t$}
\end{figure}

\subsection{Training Objective}

\begin{figure}[htbp]
\centerline{\includegraphics[scale=0.65]{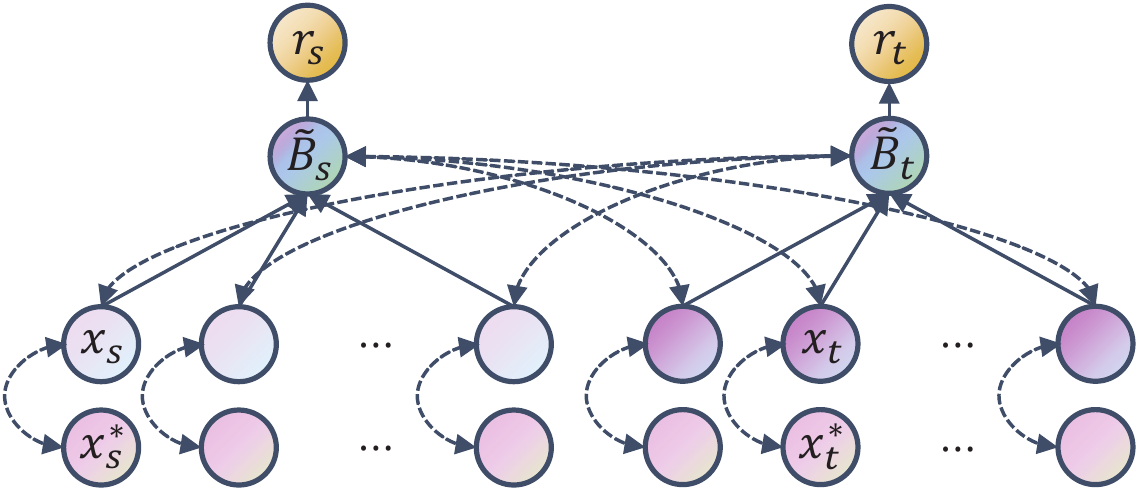}}
\caption{Contrastive Instance Learning}
\label{cil figure}
\end{figure}

As discussed above, for any instance $x_s$ in the source bag $B_s$:
(1) The instance $x_s^*$ after controllable data augmentation based on $x_s$ is its positive pair instance.
(2) The relational triple representations $\widetilde B_t$ of other different ($t\neq s$) bags in the batch are its negative pair instances.
The overall schematic diagram of \our is shown in Figure~\ref{cil figure}.

And we define the objective for instance $x_s$ in bag $B_s$ using InfoNCE~\citep{oord2018representation} loss:
\begin{equation*}
    \mathcal{L_C}(x_s;\theta) = -\log \frac{e^{\text{sim}(h_s,h_s^*)}}{e^{\text{sim}(h_s,h_s^*)}+\sum_{t:t\neq s} e^{\text{sim}(h_s,\widetilde B_t)}}
\end{equation*}
where $\text{sim}(a,b)$ is the function to measure the similarity between two representation vectors $a,b$, and $h_s=\mathcal{H}(x_s),h_s^*=\mathcal{H}(x_s^*)$ are the sentence representations of instances $x_s,x_s^*$.

Besides, to inherit the ability of language understanding from BERT and avoid catastrophic forgetting~\citep{mccloskey1989catastrophic}, we also add the masked language modeling (MLM) objective to our framework.
Pre-text task MLM randomly masks some tokens in the inputs and allows the model to predict the masked tokens, which prompts the model to capture rich semantic information in the contexts.
And we denote this objective as $\mathcal{L_M}(\theta)$.

Accordingly, the total training objective of our contrastive instance learning framework is:
\begin{equation*}
    \mathcal{L}(\theta) = \frac{\lambda(t)}{N} \sum_B \sum_{x \in B} \mathcal{L_C}(x;\theta)+\mathcal{L_B}(\theta) + \mathcal{\lambda_M} \mathcal{L_M}(\theta)
\end{equation*}
where $N=KG$ is the total number of instances in the batch, $\mathcal{\lambda_M}$ is the weight of language model objective $\mathcal{L_M}$, and $\lambda(t) \subset [0,1]$ is an increasing function related to the relative training steps $t$:
\begin{equation*}
    \lambda(t) = \frac{2}{1+e^{-t}}-1
\end{equation*}

At the beginning of our training, the value of $\lambda(t)$ is relatively small, and our framework \our focuses on obtaining an accurate bag encoder ($\mathcal{L_B}$). 
The value of $\lambda(t)$ gradually increases to 1 as the relative training steps $t$ increases, and more attention is paid to the contrastive instance learning ($\mathcal{L_C}$).

\section{Experiments}

Our experiments are designed to verify the effectiveness of our proposed framework \our.

\subsection{Benchmarks}

We evaluate our method on three popular DSRE benchmarks --- NYT10, GDS and KBP, and the dataset statistics are listed in Table~\ref{dataset statistics}.

\paragraph{NYT10}~\citep{riedel2010modeling} aligns Freebase entity relations with New York Times corpus, and it has two test set versions: (1) NYT10-D employs held-out KB facts as the test set and is still under distantly supervised. (2) NYT10-H is constructed manually by~\citep{hoffmann2011knowledge}, which contains 395 sentences with human annotations.
\paragraph{GDS}~\citep{jat2018improving} is created by extending the Google RE corpus with additional instances for each entity pair, and this dataset assures that the at-least-one assumption of MIL always holds.
\paragraph{KBP}~\citep{ling2012fine} uses Wikipedia articles annotated with Freebase entries as the training set, and employs manually-annotated sentences from 2013 KBP slot filling assessment results~\citep{ellis2012linguistic} as the extra test set.

\begin{table}[t]
    \centering
    \scalebox{0.83}{
    \begin{tabular}{l|c|c|c|c}
    \hline
    Dataset & \# Rel. & \# Ins. & \# Test Ins. & \# Test Set \\ \hline \hline
    NYT10-D & 53      & 694,491 & 172,448      & DS          \\
    NYT10-H & 25      & 362,691 & 3,777        & MA          \\
    GDS     & 5       & 18,824  & 5,663        & Partly MA   \\
    KBP     & 7       & 148,666 & 1,940        & MA          \\ \hline
    \end{tabular}
    }
    \caption{
    \label{dataset statistics}
    Statistics of various used datasets. Rel.: relation, Ins.: instance and MA: manually annotated.
    }
\end{table}

\subsection{Evaluation Metrics}

Following previous literature~\citep{lin2016neural,vashishth2018reside,alt2019fine}, we first conduct a held-out evaluation to measure model performances approximately on NYT10-D and GDS.
Besides, we also conduct an evaluation on two human-annotated datasets (NYT10-H \& KBP) to further support our claims.
Specifically, Precision-Recall curves (PR-curve) are drawn to show the trade-off between model precision and recall, the Area Under Curve (AUC) metric is used to evaluate overall model performances, and the Precision at N (P@N) metric is also reported to consider the accuracy value for different cut-offs.

\begin{table*}
    \centering
    \begin{tabular}{l|c|c|c|c|c|c|c|c|c|c}
    \hline
    \multirow{2}{*}{Method} & \multicolumn{5}{c|}{NYT10-D}                                                   & \multicolumn{5}{c}{GDS}                                                      \\ \cline{2-11} 
                            & \small{AUC}   & \small{P@100} & \small{P@200} & \small{P@300} & \small{P@M} & \small{AUC}   & \small{P@500} & \small{P@1000} & \small{P@2000} & \small{P@M} \\ \hline \hline
    Mintz$^\dagger$         & 10.7          & 52.3          & 50.2          & 45.0          & 49.2           & -             & -             & -             & -             & -              \\
    PCNN-ATT$^\ddagger$      & 34.1          & 73.0          & 68.0          & 67.3          & 69.4           & 79.9          & 90.6          & 87.6          & 75.2          & 84.5           \\
    MTB-MIL                & 40.8          & 76.2          & 71.1          & 69.4          & 72.2           & 88.5          & 94.8          & 92.2          & 87.0          & 91.3           \\
    RESIDE$^\ddagger$        & 41.5          & \underline{81.8}          & \underline{75.4}          & \underline{74.3}          & \underline{77.2}           & 89.1          & 94.8          & 91.1          & 82.7          & 89.5           \\
    REDSandT$^\ddagger$      & \underline{42.4}          & 78.0          & 75.0          & 73.0          & 75.3           & 86.1             & 95.6             & 92.6             & 84.6             & 91.0              \\
    DISTRE$^\dagger$        & 42.2          & 68.0          & 67.0          & 65.3          & 66.8           & \underline{89.9}          & \underline{97.0}          & \underline{93.8}          & \underline{87.6}          & \underline{92.8}           \\ \hline
    CIL$^*$                     & \textbf{50.8} & \textbf{90.1} & \textbf{86.1} & \textbf{81.8} & \textbf{86.0}  & \textbf{91.6} & \textbf{98.4} & \textbf{95.3} & \textbf{88.7} & \textbf{94.1}  \\ \hline
    \end{tabular}
    \caption{
    \label{overall performance table}
    Model performances on NYT10-D and GDS. ($\dag$)$/$($\ddag$) marks the results on (NYT10-D column)$/$(both columns) are reported in the previous papers. Bold and underline indicate the best and the second best scores, and $*$ indicates that our model shows significant gains ($p>0.05$) over the second-best model based on Student's t-test.
    }
\end{table*}

\subsection{Baseline Models}

We choose six recent methods as baseline models.
\paragraph{Mintz}~\citep{mintz2009distant} A multi-class logistic regression RE model under DS setting.
\paragraph{PCNN-ATT}~\citep{lin2016neural} A piece-wise CNN model with selective attention over instances.
\paragraph{MTB-MIL}~\citep{baldini-soares-etal-2019-matching} A relation learning method based on distributional similarity, achieves amazing results for supervised RE\footnote{For MTB-MIL, we firstly conduct MTB pre-training to learn relation representations on the entire training corpus and continually fine-tune the model by the MIL framework.}.
\paragraph{RESIDE}~\citep{vashishth2018reside} A NN model that makes use of relevant side information (entity types and relational phrases) and employs Graph CNN to capture syntactic information of instances.
\paragraph{REDSandT}~\citep{DBLP:journals/corr/abs-2102-01156} A transformer-based DSRE method that manages to capture highly informative instance and label embeddings by exploiting BERT pre-trained model.
\paragraph{DISTRE}~\citep{alt2019fine} A transformer-based model, GPT fine-tuned for DSRE under the MIL.

\subsection{Evaluation on Distantly Supervised Set}

We summarize the model performances of our method and above-mentioned baseline models in Table~\ref{overall performance table}.
From the results, we can observe that:
(1) On both two datasets, our proposed framework \our achieves the best performance in all metrics.
(2) On NYT10-D, compared with the previous SOTA model DISTRE, \our improves the metric AUC (42.2$\to$50.8) by 20.4\% and the metric P@Mean (66.8$\to$86.0) by 28.7\%.
(3) On GDS, though the metric of previous models is already high ($\approx 90.0$), our model still improves it by nearly 2 percentage points. (89.9$\to$91.6 \& 92.8$\to$94.1).

\begin{figure}[htbp]
\centerline{\includegraphics[width=1.05\linewidth]{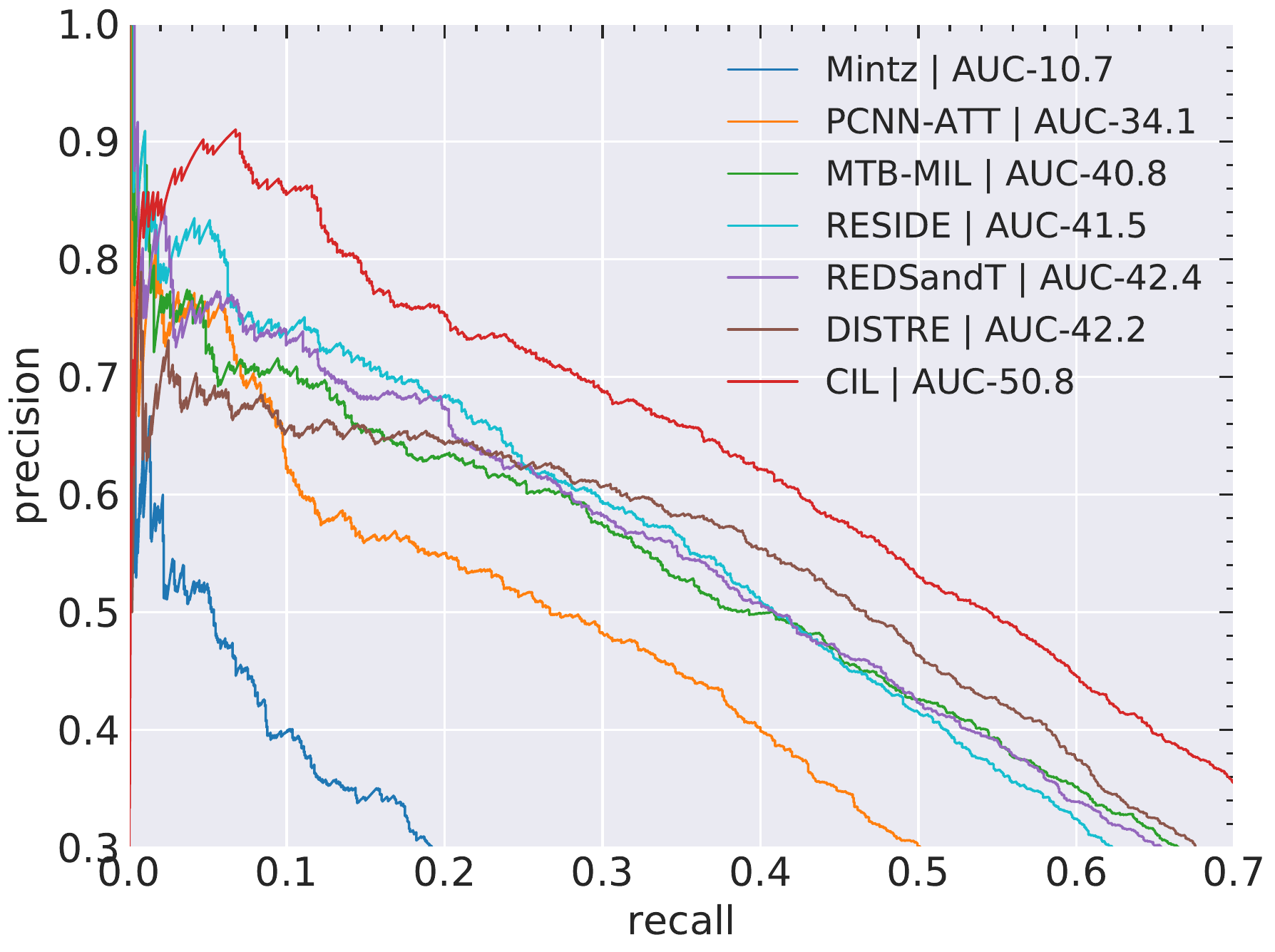}}
\caption{PR-curve on NYT10-D.}
\label{Overall PR-Curve}
\end{figure}

The overall PR-curve on NYT10-D is visualized in Figure~\ref{Overall PR-Curve}.
From the curve, we can observe that:
(1) Compared to PR-curves of other baseline models, our method shifts up the curve a lot. 
(2) Previous SOTA model DISTRE performs worse than model RESIDE at the beginning of the curve and yields a better performance after a recall-level of approximately 0.25, and our method \our surpasses previous two SOTA models in all ranges along the curve, and it is more balanced between precision and recall.
(3) Furthermore, as a SOTA scheme of relation learning, MTB fails to achieve competitive results for DSRE. This is because MTB relies on label information for pre-training, and noisy labels in DSRE may influence its model performance.

\subsection{Evaluation on Manually Annotated Set}

The automated held-out evaluation may not reflect the actual performance of DSRE models, as it gives false positive/negative labels and incomplete KB information.
Thus, to further support our claims, we also evaluate our method on two human-annotated datasets, and the results\footnote{Manual evaluation is performed for each test sentence.} are listed in Table~\ref{human evaluation}.

\begin{table}[htbp]
    \centering
    \scalebox{0.88}{
    \begin{tabular}{l|c|c|c|c|c|c}
    \hline
    \multirow{2}{*}{Method} & \multicolumn{3}{c|}{NYT10-H}                   & \multicolumn{3}{c}{KBP}                        \\ \cline{2-7} 
                            & \small{AUC}   & \small{F1}    & \small{P@M} & \small{AUC}   & \small{F1}    & \small{P@M} \\ \hline \hline
    PCNN-A                  & \underline{38.9}          & 47.0          & \underline{58.6}           & 15.4          & 31.5          & 32.8           \\
    DISTRE                  & 37.8          & \underline{50.9}          & 54.1           & \underline{22.1}          & \underline{37.5}          & \underline{46.4}           \\ \hline
    CIL                     & \textbf{46.0} & \textbf{55.5} & \textbf{63.0}  & \textbf{30.1} & \textbf{44.0} & \textbf{48.2}  \\ \hline
    \end{tabular}
    }
    \caption{
    \label{human evaluation}
    Model performances on NTY10-H and KBP. PCNN-A denotes PCNN-ATT. F1 refers to Micro-F1.
    }
\end{table}

From the above result table, we can see that:
(1) Our proposed framework \our can still perform well under accurate human evaluation, with averagely 21.7\% AUC improvement on NYT10-H and 36.2\% on KBP, which means our method can generalize to real scenarios well.
(2) On NYT10-H, DISTRE fails to surpass PCNN-ATT in metric P@Mean.
This indicates that DISTRE gives a high recall but a low precision, but our method \our can boost the model precision (54.1$\to$63.0) while continuously improving the model recall (37.8$\to$46.0).
And the human evaluation results further confirm the observations in the held-out evaluation described above.

\begin{figure}[htbp]
\centerline{\includegraphics[width=1.05\linewidth]{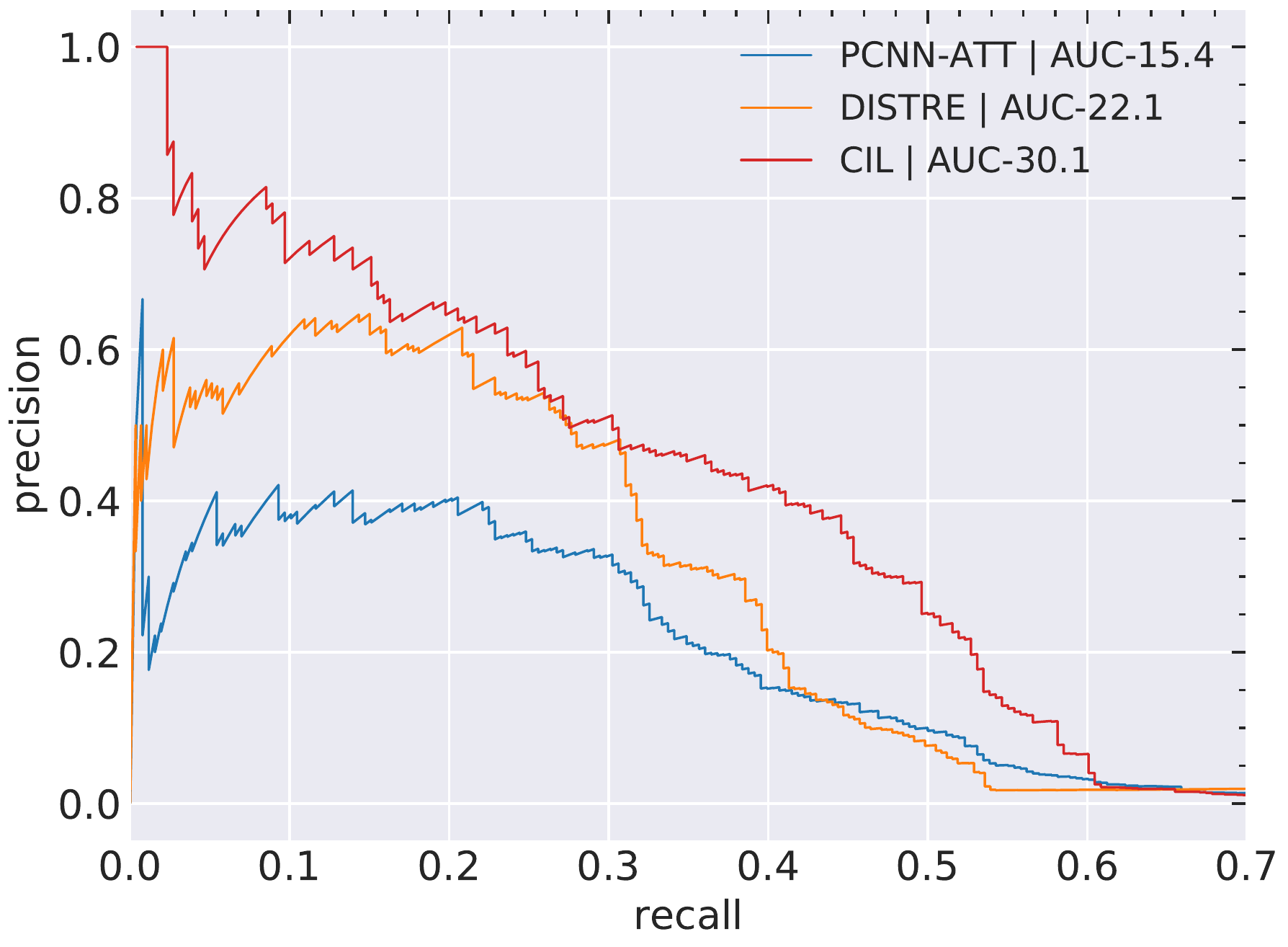}}
\caption{PR-Curve on KBP.}
\label{KBP PR-Curve}
\end{figure}

We also present the PR-curve on KBP in Figure~\ref{KBP PR-Curve}.
Under accurate sentence-level evaluation on KBP, the advantage of our model is more obvious with averagely 36.2\% improvement on AUC, 17.3\% on F1 and 3.9\% on P@Mean, respectively.

\subsection{Ablation Study}

To further understand our proposed framework \our, we also conduct ablation studies.

We firstly conduct an ablation experiment to verify that \our has utilized abundant instances inside bags:
(1) By removing our proposed contrastive instance learning, the framework degenerates into vanilla MIL framework, and we train the MIL on regular bags (MIL$_{bag}$).
(2) To prove the MIL can not make full use of sentences, we also train the MIL on sentence bags (MIL$_{sent}$), which repeats each sentence in the training corpus to form a bag\footnote{All the results in Table~\ref{training pattern} are obtained under the same test setting that uses MIL bags~(\ie BERT+ATT) as test units.}.

\begin{table}[htbp]
\centering
\scalebox{0.95}{
\begin{tabular}{l|c|c|c}
\hline
Method & AUC           & F1            & P@M        \\ \hline \hline
CIL    & \textbf{50.8} & \textbf{52.2} & \textbf{86.0} \\
MIL$_{bag}$    & 40.3(-10.5)   & 47.1(-5.1)    & 70.0(-16.0)   \\
MIL$_{sent}$    & 36.0(-14.8)   & 43.5(-8.7)    & 63.3(-22.7)   \\ \hline
\end{tabular}}
\caption{Model performances of three training patterns.}
\label{training pattern}
\end{table}

From Table~\ref{training pattern} we can see that:
(1) MIL$_{bag}$ only resorts to the accurate bag-level representations to train the model and fails to play the role of each instance inside bags; thus, it performs worse than our method \our (50.8$\to$40.3).
(2) Though MIL$_{sent}$ can access all training sentences, it loses the advantages of noise reduction in MIL$_{bag}$~(40.3$\to$30.6). The noisy label supervision may wrongly guide model training, and its model performance heavily suffers from DS data noise~(86.0$\to$63.3).
(3) Our framework \our succeeds in leveraging abundant instances while retaining the ability to denoise.

To validate the rationality of our proposed positive/negative pair construction strategy, we also conduct an ablation study on three variants of our framework \our.
We denote these variants as:

\noindent \textbf{CIL$_{randpos}$}: Randomly select an instance $x_{s'}$ also from bag $B_s$ as the positive pair instance for $x_s$.

\noindent\textbf{CIL$_{bagpos}$}: Just take the relational triple representation $\widetilde B_s$ as the positive pair instance for $x_s$.

\noindent \textbf{CIL$_{randneg}$}: Randomly select an instance $x_{t}$ from another bag $B_t$ as the negative pair instance for $x_s$.

And we summarize the model performances of our \our and other three variants in Table~\ref{ablation table}.

\begin{table}[htbp]
\centering
\scalebox{0.95}{
\begin{tabular}{l|c|c|c}
\hline
Method          & AUC           & F1            & P@M        \\ \hline \hline
CIL    & \textbf{50.8} & \textbf{52.2} & \textbf{86.0} \\
CIL$_{randpos}$ & 49.2(-1.6)    & 50.9(-1.3)    & 83.8(-2.2)    \\
CIL$_{bagpos}$  & 47.8(-3.0)    & 50.5(-1.7)    & 79.2(-6.8)    \\
CIL$_{randneg}$ & 48.4(-2.4)    & 50.6(-1.6)    & 78.2(-7.8)    \\ \hline
\end{tabular}}
\caption{
\label{ablation table}Model performances of our proposed framework \our and its three variants.
}
\end{table}

As the previous analysis in section~\ref{sec: cil}, the three variants of our \our framework may suffer from DS noise: 
(1) Both variants CIL$_{randpos}$ and CIL$_{bagpos}$ may construct noisy positive pairs; therefore, their model performances have a little drop (50.8$\to$49.2, 50.8$\to$47.8). Besides, the variant CIL$_{bagpos}$ also relies on the bag encoder, for which it performs worse than the variant CIL$_{randpos}$ (49.2$\to$47.8). 
(2) Though the constructed negative pairs need not be as accurate as positive pairs, the variant CIL$_{randneg}$ treats all instances equally, which gives up the advantage of formed accurate representations. Thus, its model performance also declines (50.8$\to$48.4).

\subsection{Case Study}

We select a typical bag (see Table~\ref{case bag}) from the training set to better illustrate the difference between MIL$_{sent}$, MIL$_{bag}$ and our framework \our.

\begin{table}[htbp]
\centering
\scalebox{0.90}{
\begin{tabular}{l|c}
\hline
Sentence                                                                                                                                         & Predicted Relation                                                                                                                   \\ \hline \hline
\begin{tabular}[c]{@{}c@{}}\textit{john mcgahern}, the eldest \\ of seven children, was born\\ on nov.12, 1934, in \textit{dublin}.\end{tabular} & \begin{tabular}[c]{@{}c@{}}\phantom{C}\llap{S}: \textit{/place\_borned} \ding{51}\\ \phantom{C}\llap{B}: \textit{/place\_borned} \ding{51}\\ C: \textit{/place\_borned} \ding{51}\end{tabular} \\ \hline
\begin{tabular}[c]{@{}c@{}}\textit{john mcgahern}, whose stark\\ ..., died yesterday in \textit{dublin}.\end{tabular}                            & \begin{tabular}[c]{@{}l@{}}\phantom{C}\llap{S}: \textit{/place\_borned} \ding{55}\\ \phantom{C}\llap{B}: \textit{/place\_borned} \ding{55}\\ C: \textit{/place\_deaded} \ding{51}\end{tabular} \\ \hline
\end{tabular}}
\caption{A typical bag selected from the training set: The bag is constructed with relational triple (\textit{john mcgahern}, \textit{/place\_borned}, \textit{dubin}), and the first sentence (S1) is clean to express relation \textit{/place\_borned} while the second instance (S2) are noisy with true relation \textit{/place\_deaded}. S: MIL$_{sent}$, B: MIL$_{bag}$ and C: CIL.}
\label{case bag}
\end{table}

Under MIL$_{sent}$ pattern, both S1, S2 are used for model training, and the noisy sentence S2 may confuse the model.
As for MIL$_{bag}$ pattern, S1 is assigned with a high attention score while S2 has a low attention score. However, MIL$_{bag}$ only relies on the bag-level representations, and sentences like S2 can not be used efficiently.
Our framework \our makes full use of all instances (S1, S2) and avoids the negative effect of DS data noise from S2.

\section{Related Work}

Our work is related to DSRE, pre-trained language models, and recent contrastive learning methods.

\paragraph{DSRE}

Traditional supervised RE systems heavily rely on the large-scale human-annotated dataset, which is quite expensive and time-consuming.
Distant supervision is then introduced to the RE field, and it aligns training corpus with KB facts to generate data automatically.
However, such a heuristic process results in data noise and causes classical supervised RE models hard to train.
To solve this issue,~\citet{lin2016neural} applied the multi-instance learning framework with selective attention mechanism over all instances, and it helps RE models learn under DS data noise.
Following the MIL framework, recent works improve DSRE models from many different aspects:
(1)~\citet{yuan2019cross} adopted relation-aware attention and constructed super bags to alleviate the problem of bag label error.
(2)~\citet{ye2019looking} analyzed the label distribution of dataset and found the shifted label problem that significantly influences the performance of DSRE models.
(3)~\citet{vashishth2018reside} employed Graph Convolution Networks~\citep{defferrard2016convolutional} to encode syntactic information from the text and improves DSRE models with additional side information from KBs.
(4)~\citet{alt2019fine} extended the GPT to the DSRE, and fine-tuned it to achieve SOTA model performance.

\paragraph{Pre-trained LM}

Recently pre-trained language models achieved great success in the NLP field.
~\citet{vaswani2017attention} proposed a self-attention based architecture --- Transformer, and it soon becomes the backbone of many following LMs.
By pre-training on a large-scale corpus, BERT~\citep{devlin-etal-2019-bert} obtains the ability to capture a notable amount of “common-sense” knowledge and gains significant improvements on many tasks following the fine-tune scheme. 
At the same time, GPT~\citep{radford2018improving}, XL-Net~\citep{yang2019xlnet} and GPT2~\citep{radford2019language} are also well-known pre-trained representatives with excellent transfer learning ability.
Moreover, some works~\citep{radford2019language} found that considerably increasing the size of LM results in even better generalization to downstream tasks.

\paragraph{Contrastive Learning}

As a popular unsupervised method, contrastive learning aims to learn representations by contrasting positive pairs against negative pairs~\citep{hadsell2006dimensionality,oord2018representation,chen2020simple,he2020momentum}.
\citet{wu2018unsupervised} proposed to use the non-parametric instance-level discrimination to leverage more information in the data samples.
Our approach, however, achieves the goal of data-efficiency in a more complicated MIL setting: instead of contrasting the instance-level information during training, we find that instance-bag negative pair is the most effective method, which constitutes one of our main contributions.
In the NLP field, ~\citet{dai2017contrastive} proposed to use contrastive learning for image caption, and~\citet{clark2020electra} trained a discriminative model for language representation learning.
Recent literature~\citep{peng-etal-2020-learning} has also attempted to relate the contrastive pre-training scheme to classical supervised RE task.
Different from our work,~\citet{peng-etal-2020-learning} aims to utilize abundant DS data and help classical supervised RE models learn a better relation representation, while our \our focuses on learning an effective and efficient DSRE model under DS data noise.

\section{Conclusion}

In this work, we discuss the long-standing DSRE framework (\ie MIL) and argue the MIL is not efficient enough, as it aims to form accurate bag-level representations but sacrifices the potential information of abundant instances inside MIL bags.
Thus, we propose a contrastive instance learning method \our to boost the MIL model performances.
Experiments have shown the effectiveness of our \our with stable and significant improvements over several baseline models, including current SOTA systems.

\section*{Acknowledgments}

This work has been supported in part by National Key Research and Development Program of China (2018AAA0101900), Zhejiang NSF (LR21F020004), Key Technologies and Systems of Humanoid Intelligence based on Big Data (Phase ii) (2018YFB1005100), Zhejiang University iFLYTEK Joint Research Center, Funds from City Cloud Technology (China) Co. Ltd., Zhejiang University-Tongdun Technology Joint Laboratory of Artificial Intelligence, Chinese Knowledge Center of Engineering Science and Technology (CKCEST).

\bibliographystyle{acl_natbib}
\bibliography{acl2021}

\end{document}